\documentclass{ISMA_USD2020}

\usepackage{graphicx}
\usepackage{amsmath}
\usepackage{float}
\usepackage{amsfonts}
\usepackage{algorithm}
\usepackage{algorithmicx}
\usepackage{algpseudocode}
\usepackage{epstopdf}
\usepackage{bm}
\usepackage{tikz}
\usepackage{multirow}
\usepackage{amsmath}
\usetikzlibrary{bayesnet}
\usetikzlibrary{shapes.gates.logic.US,trees,positioning,arrows}
\usepackage{caption}
\usepackage{subcaption}
\usetikzlibrary{trees}
\usepackage{graphicx}
\usepackage{mathtools}
\usepackage{amsmath}
\usepackage{booktabs}
\usepackage{rotating}
\usepackage{booktabs,makecell}
\usepackage{hyphenat}
\usepackage{cite}

\usepackage{amsthm}

\usepackage{epstopdf}
\usepackage{amssymb}
\usepackage{microtype}
\usepackage{url}
\usepackage{caption}
\usepackage{subcaption}
\usepackage{tabularray}
	
\hypersetup{pdftitle  = {Cost-informed dimensionality reduction},
	pdfauthor = {A.J.\ Hughes, K.\ Worden, N.\ Dervilis, T.J.\ Rogers},
	pdfkeywords = {dimensionality reduction, classification, decision-making, digital twins}}

\title{\sloppy{\nohyphens{Cost-informed dimensionality reduction for structural digital twin technologies}}}

\author[1,$\ast$] {{A.\ J.\ Hughes}}
\author[1,2] {K.\ Worden}
\author[1] {N.\ Dervilis}
\author[1] {T.\ J.\ Rogers}

\affil[1] {Dynamics Research Group,  Department of Mechanical Engineering, University of Sheffield,  \NewLineAffil
			Mappin Street, Sheffield, S1 3JD, UK}
		
\affil[2] {The Alan Turing Institute, The British Library, 96 Euston Road, London, NW1 2DB, UK  \NewLineAffil $^{\ast}$e-mail: \textbf{aidan.j.hughes@sheffield.ac.uk} \NewLineAffil}
		
\date{June 2022}

\begin{document}



\abstract{\\  Classification models are a key component of structural digital twin technologies used for supporting asset management decision-making. An important consideration when developing classification models is the dimensionality of the input, or feature space, used. If the dimensionality is too high, then the `curse of dimensionality' may rear its ugly head; manifesting as reduced predictive performance. To mitigate such effects, practitioners can employ dimensionality reduction techniques. The current paper formulates a decision-theoretic approach to dimensionality reduction for structural asset management. In this approach, the aim is to keep incurred misclassification costs to a minimum as the dimensionality is reduced and discriminatory information may be lost. This formulation is constructed as a eigenvalue problem, with separabilities between classes weighted according to the cost of misclassifying them when considered in the context of a decision process. The approach is demonstrated using a synthetic case study.} 

\maketitle


\section{Introduction}

Decision-support technologies such as digital twins are becoming increasingly important for the management of structures; to maximise the remaining useful life and to minimise the environmental cost of existing infrastructure \cite{Grieves2017,Gardner2020digitwin,Kapteyn2021probabilistic,Worden2020digital}. One of the core technologies underpinning online asset-management tools such as digital twins and health monitoring systems are statistical classifiers –- used to identify and characterise discrete latent states pertinent to decision-making such as damage, operational, and environmental conditions\footnote{Here, it is worth stating that the current paper uses `decision' to refer to the selection of an action that is either observational (e.g. inspection), or interventional (e.g. repair). This usage is somewhat different to literature in the field of machine learning that uses `decision' to refer to the selection of a class label for a given data point.}.

A key consideration when developing classification models is the dimensionality of the input, or feature, space in which the data used to discriminate between classes exists. High-dimensional feature spaces can bring with them increased computational complexity \cite{Cunningham2021knn}; requiring more resource or time to conduct analyses. The trade-off between required computational resource and computation time is a critical design consideration in the development of data-based decision-support tools, as it governs the rate at which predictions and decisions can be made online, and the economic viability of such systems.  More generally, the `curse of dimensionality' is well-studied in the fields of statistical machine learning \cite{Bellman1966dynamic,Bishop1995neural,Bengio2005curse,Bishop2006}. In the context of machine learning, the primary way in which the `curse of dimensionality' manifests is as a degradation in generalisation performance when increasing the dimensionality of a feature space beyond a certain point \cite{Trunk1979problem, Mclachlan2005discriminant}. In the context of structural asset management, it is desirable to avoid such degradation in performance as misclassifications translate to incurred costs in terms of resources and/or safety.

 To mitigate the adverse effects associated with high-dimensional feature spaces --- and to ensure online decisions can be made in a timely manner, while adhering to constraints regarding the required computational resource --- practitioners can employ dimensionality reduction techniques when developing statistical models. Dimensionality reduction techniques seek to find a reduced set of features that preserve some useful properties of the higher-dimensional data. Many dimensionality reduction methods used commonly in structural dynamics, such as principal component analysis (PCA), find subspaces on a purely information-theoretic basis; in the case of PCA, a subspace is found that retains the maximum possible variance in the data \cite{Wold1987principal}. While broadly applicable in many situations, such approaches do not take into account the fact that, in order to minimise the rate of misclassification, class separability should be preserved rather than, say, variance \cite{Fukunaga1990introduction,Vapnik1991principles}. A common dimension reduction technique that preserves class separability is linear discriminant analysis (LDA) \cite{Mclachlan2005discriminant}. Although LDA maximises class separability and thus minimises the overall rate of misclassification, classes are separated with equal priority and thus assumes that all types of misclassification are equally detrimental -- when considered in the context of decision-making for engineering structures, this assumption of LDA is not necessarily optimal. 

In engineering applications, there are often non-uniform and asymmetric costs associated with misclassifications. A classic example of this arises in structural health monitoring (SHM) where false-positives (type-I errors), in which structures are classified as being in a damaged state when in fact they are undamaged, result in unnecessary actions being taken (e.g inspections) which will have some monetary cost associated. Similarly, false-negatives (type-II) errors, in which structures are mistakenly classified as being undamaged, result in erroneous inaction - potentially leading to catastrophic failure of structures, which in turn will have some large cost associated.

In the field of structural asset management, there has been limited investigation into dimensionality reduction techniques that account for non-uniform and asymmetric misclassification costs; however, cost-sensitive discriminant analysis has been studied in the context of computer vision for face recognition in \cite{Lu2012cost,Lu2013cost,Li2020cost}.

 The current work seeks to develop an algorithm for conducting dimensionality reduction that explicitly considers the decision process in which a classifier is employed, by selecting dimensions in a subspace that prioritise maintaining separability between pairs of classes that have high misclassification costs. This property is achieved by adapting multi-class LDA to include a cost-based weighting scheme. The proposed cost-informed dimensionality reduction approach is compared to both PCA and standard LDA using a case study based on a synthetic dataset, designed to aid in visualising the differences between candidate approaches. While the proposed cost-informed approach is somewhat rudimentary, the results of the case study provide strong motivation for the further development of decision-theoretic dimensionality reduction methods, in addition to the development of decision-theoretic signal processing techniques more generally, with applications in structural digital twin technologies.
 
 The layout of the current paper is as follows. Section \ref{sec:background} provides background theory on classification, PCA, and LDA. Section \ref{sec:CIDR} introduces the cost-informed dimensionality reduction approach. Section \ref{sec:CaseStudy} demonstrates the proposed dimension reduction approach using a visual case study. Finally, Sections \ref{sec:disc} and \ref{sec:conc} provided further discussion and conclusions, respectively.

\section{Background Theory}\label{sec:background}

\subsection{Statistical Classification and Decision-making}

In general, classification is the task associated with categorising observations $\mathbf{x} \in X$ according to discrete labels $y \in Y$, where $X$ and $Y$ denote the \textit{input/feature space} and \textit{label space}, respectively. A probabilistic perspective can be adopted by defining inputs as $D$-dimensional random vectors, such that $X = \mathbb{R}^D$, and descriptive labels as discrete random variables, such that $Y = \{1, \ldots, K\}$, where $K$ is the number of target classes. In learning a statistical classifier, one wishes to obtain a robust mapping from the input space to the label space, i.e.\ $f : X \rightarrow Y$ -- for a given set of observed features $\mathbf{x}$, a probabilistic classifier $f$ provides a categorical probability distribution $p(y|\mathbf{x})$ over the label space. Adopting a probabilistic approach to classification is often useful for decision-making applications as it allows one to account for uncertainty within the decision-making process, thereby providing some degree of robustness; nonetheless, a label can be assigned corresponding to the class with the highest probability, i.e.\ $\hat{y} = \text{argmax}_{y}p(y|\mathbf{x})$.

For classification to be effective, data associated with disparate classes should be separable in the feature space, i.e.\ classes do not overlap. If, for a given dataset, each class occupies its own disjoint region of the feature space, then perfect classification is achievable; however, in practice, this condition is seldom met for a variety of reasons including noise processes affecting data, and varying degrees of feature sensitivity. As a result, misclassifications arise in which incorrect labels are attributed to data. It follows that, from an information-theoretic perspective, learning an effective classifier amounts to minimising the rate of misclassifications for unseen data -- this notion is captured by empirical risk minimisation \cite{Vapnik1991principles} and maximal margin classifiers such as the support vector machine \cite{Cortes1995}. In the probabilistic setting, regions of overlap between classes correspond to a high degree of uncertainty in label predictions, i.e.\ the predictive distributions $p(y|\mathbf{x})$ have high entropy \cite{Shannon1948mathematical}.

In the context of structural digital twin technologies, statistical classifiers are useful as they provide a data-based framework for predicting latent states pertinent to decision-making that are otherwise impractical or impossible to measure directly. Such states of interest may pertain to structural damage; for example, damage types, extents, or locations \cite{Farrar2013}. Additionally, states might be related to other operational or environmental conditions such as temperature regions, or loading modes. The predictive distribution $p(y|\mathbf{x})$ obtained from a statistical classifier, which encodes one's beliefs about the salient latent states, can be used in a decision-theoretic framework to determine suitable courses of action such as maintenance strategies \cite{Hughes2021} or control actions \cite{Guzman2022adaptive}.

As mentioned previously, poor predictions or misclassifications can result in the selection of suboptimal actions and strategies -- by definition, such courses of actions incur costs that would otherwise have been avoided had the optimal actions have been selected. If one considers the possible costs that can be attributed to misclassifications in the context of decision-making using structural digital twins, one can readily arrive at the conclusion that these misclassification costs are, in general, \textit{non-uniform} (i.e. the cost associated with misclassifying label $a$ as label $b$ differs from the cost associated with misclassifying label $a$ as label $c$) and \textit{asymmetric} (i.e. the cost associated with misclassifying label $a$ as label $b$ differs from the cost associated with misclassifying label $b$ as label $a$). One example of where these asymmetry and non-uniformity properties arise is in SHM. In a SHM maintenance decision process, misclassifying a benign undamaged state as an advanced damage state would result in some unnecessary inspection or maintenance action being performed with a corresponding cost; misclassifying an advanced damaged state as a benign undamaged state would result in some critical intervention being missed potentially leading to the catastrophic failure of the system and thus incurring a higher cost. Additionally, misclassifying two benign undamaged states (e.g.\ room and cold temperature) is less costly than misclassifying the benign as undamaged, as two different benign states typically warrant the same, or very similar, courses of action.

The task of learning an effective classifier can be aided by feature selection, and/or feature projection techniques, to produce input spaces with high separability between classes. This feature selection/projection process can often be used for an additional purpose --- dimensionality reduction.

\subsection{Dimensionality Reduction}

Dimensionality reduction techniques are a well-studied topic in machine learning and data science \cite{Van2009dimensionality,Reddy2020analysis}. The techniques have received much attention, in part, due to the aforementioned `curse of dimensionality'. In the context of machine learning and classification, the `curse of dimensionality' typically manifests as a degradation in predictive performance as dimensionality increases beyond a certain point -- this is sometimes referred to as the \textit{peaking phenomenon} as the predictive performance also degrades for very low numbers of dimensions as discriminative information is lost. For reasons previously discussed, degradation in predictive performance is less than ideal in the context of decision-making for structural asset management as an increase in misclassifications results in an increase in costs incurred. An additional consideration associated with the `curse of dimensionality' relates to the data requirements for learning classification  models in high-dimensional feature spaces; in general, higher-dimensional data require a greater amount of data to model \cite{Silverman1998density,Koutroumbas2008pattern}. This is a highly relevant consideration for structural digital twins as data from large-scale and high-value systems are often scarce and/or prohibitively expensive to obtain, particularly for applications where damage-state data are required. Finally, it is worth reiterating that high-dimensional feature spaces can result in increased computational cost for classification; for example, a $K$-nearest-neighbours classifier (based on the Minkowski distance) has a prediction time complexity of $\mathcal{O}(nD)$ where $n$ is the number of data samples \cite{Cunningham2021knn}. The computational complexity is pertinent to decision support systems such as digital twins because it determines, for a given amount of computational resource, the rate at which online decisions can be made. This criterion is a critical consideration in the digital twin design and development process as the timescales over which phenomena of interest can occur in engineering systems vary from decades down to nanoseconds.

In general, dimensionality reduction techniques seek to find an embedding of features $\mathbf{x} \in \mathbb{R}^D$ as transformed features $\mathbf{x}^{\prime} \in \mathbb{R}^d$ such that $d < D$ while retaining some desirable properties of the data. Approaches to dimensionality reduction can broadly be categorised as linear, or nonlinear. As one would expect, linear reduction techniques are accomplished through affine transformations. Nonlinear approaches have developed using methods such as kernel projection \cite{Scholkopf1998nonlinear} and topological analysis \cite{Mcinnes2018umap}. Furthermore, approaches can be categorised as feature selection techniques and feature projection techniques. In feature selection approaches, reduced feature sets are typically obtained by selecting a subset of the original features. Projection-based approaches yield lower-dimensional subspaces using reduced-rank transformations.

The remainder of this section focusses on linear data projection approaches and provides an overview of two commonly-used dimensionality reduction techniques -- PCA and LDA.

\subsubsection{Principal Component Analysis}

First introduced at the turn of the twentieth century, PCA is one of the most ubiquitous approaches to dimensionality reduction \cite{Pearson1901pca}. PCA seeks to find a subspace in which, for a given dimensionality $d$, maximal variance in the data is preserved.

There are many ways to formulate PCA, one of the most convenient is to formulate it as an eigendecomposition of the empirical covariance matrix of the data. This approach is equivalent to fitting an ellipsoid to the data and finding the principle axes of said ellipsoid.

Starting with an $N \times D$ data matrix $\mathbf{X}$ comprising datapoints in the original feature space $\mathbf{x}$, one can compute the empirical covariance of the data as,

\begin{equation}\label{eq:empirical_scat}
	\Sigma_e = \mathbf{X}^{\top}\mathbf{X}.
\end{equation}

\noindent By solving an eigenvalue problem, this covariance matrix can then be decomposed as,

\begin{equation}
	\Sigma_e = V \Lambda V^{\top},
\end{equation}

\noindent where V is a $D \times D$ matrix of eigenvectors and $\Lambda$ is a diagonal matrix containing the eigenvalues.

 The $N \times D$ data matrix $\mathbf{X}$ can be transformed to a coordinate system in which the basis vectors preserve variability as follows,

\begin{equation}\label{eq:proj}
	\mathbf{X}^{\prime} = \mathbf{X}V
\end{equation}

\noindent where $\mathbf{X}^{\prime}$ is $N \times D$ matrix comprising the projected features $\mathbf{x}^{\prime}$.

Assuming eigenvectors have been ordered with respect to descending eigenvalues (for PCA, the eigenvalues are proportional to the explained variance in the direction of the corresponding eigenvector), $\mathbf{X}^{\prime}$ can then reduced to $N \times d$ while retaining as much variance as possible by sequentially discarding columns, starting with the rightmost. Alternatively, this can be accomplished by discarding columns of $V$ prior to transforming the data; in this case, the span of the remaining eigenvectors forms the hyperplane onto which the data are projected.

Here, it is worth highlighting some of the interesting properties of PCA. As the empirical covariance matrix is symmetric, the basis vectors obtained via PCA are orthogonal. Additionally, PCA can be conducted in a fully unsupervised manner as no class labels for data are required.

\subsubsection{Linear Discriminant Analysis}

In contrast to PCA, LDA seeks to find a subspace that preserves maximal separability between classes -- to accomplish this, LDA adopts a supervised learning approach, meaning class labels are required to find a transformation \cite{Mclachlan2005discriminant}. Similar to PCA, multi-class LDA can be constructed as an eigendecomposition.

To assess class separability in a given direction, one must first find the scatter within classes $\Sigma_w$, and the scatter between classes $\Sigma_b$.

LDA makes the assumption that the scatter within classes is common to all classes. Thus, the scatter within classes can be assumed to be given by the average,

\begin{equation}\label{eq:LDA1}
	\Sigma_w = \frac{1}{K} \sum_{k=1}^{K}(\mathbf{X}_{k} - \bar{\mathbf{x}}_k)^{\top}(\mathbf{X}_{k} - \bar{\mathbf{x}}_k)
\end{equation} 

\noindent where $K$ is the number of classes, $\mathbf{X}_{k}$ denotes the data in $\mathbf{X}$ associated with class $k$, and $\bar{\mathbf{x}}_k$ denotes the empirical mean (or centroid) of the data associated with class $k$. The scatter between classes is then given by,

\begin{equation}\label{eq:LDA2}
	\Sigma_b = \sum_{k=1}^{K}(\bar{\mathbf{x}}_k - \bar{\mathbf{x}})^{\top}(\bar{\mathbf{x}}_k - \bar{\mathbf{x}})
\end{equation}

\noindent where $\bar{\mathbf{x}}$ is the empirical mean for data associated with all classes. 

It follows that the separability $\psi$ in a given direction $\mathbf{u}$ can be expressed as,

\begin{equation}\label{eq:separability}
	\psi = \frac{\mathbf{u}^{\top}\Sigma_b\mathbf{u}}{\mathbf{u}^{\top}\Sigma_w\mathbf{u}}.
\end{equation}

This representation of class separability can be understood intuitively; $\psi$ increases if the data within each class become less spread out (i.e.\ $|\Sigma_w|$ decreases), or if the centroids for each class become more spread out (i.e.\ $|\Sigma_b|$ increases).

The direction that maximises class separability is given by,

\begin{equation}
	\hat{\mathbf{u}} = \underset{\mathbf{u}}{\text{argmax}}\psi
\end{equation}

As eluded to earlier, this maximisation can be accomplished by solving the eigendecomposition,

\begin{equation} \label{eq:LDA6}
	\Sigma_r = V \Lambda V^{\top}
\end{equation}

\noindent again where $\Sigma_r=\Sigma_w^{-1}\Sigma_b$, and where $\Lambda$ is a diagonal matrix of eigenvalues, and $V$ is a $D \times D$ matrix of eigenvectors. For LDA, the eigenvalues are related to the total class separability in the direction of the corresponding eigenvector.

At this stage, the process of transforming/projecting data and reducing the dimensionality is identical to that as specified for PCA and given in Equation (\ref{eq:proj}). Here, a few differences between LDA and PCA are noted. Unlike PCA, LDA does not generally yield orthogonal basis vectors, unless $\Sigma_r$ is symmetric -- this condition is met when at least one of $\Sigma_w$ or $\Sigma_b$ is a scalar multiple of the identity matrix. Moreover, whereas PCA can be conducted in an unsupervised manner, LDA requires class labels for the data and thus is a supervised approach.

By considering Equations (\ref{eq:LDA1}) to (\ref{eq:LDA6}), one can realise that separability is treated equally across all classes -- this means that LDA implicitly assumes that all misclassifications are equally concerning. For reasons previously discussed, this is not necessarily a desirable property for a dimensionality-reduction algorithm when using classification models on a reduced feature space to support decision-making. To overcome this limitation associated with LDA, a cost-informed approach is proposed in the following section.

\section{Cost-informed Dimensionality Reduction}\label{sec:CIDR}

As discussed in Section \ref{sec:background}, in the context of structural asset management, misclassification costs are often asymmetric and non-uniform. Given this fact, it would then be prudent to use dimensionality reduction techniques that account for such costs when deploying structural digital twins based on statistical classifiers.

\subsection{Cost Function}

As the target of this approach is to account for misclassification costs, one must first define the set of possible classification outcomes. Assuming each datapoint has only one corresponding label associated, and that a statistical classifier assigns a single label per datapoint, the set of classification outcomes is given by the Cartesian product $Y \times Y$. Each element of this set corresponds to a 2-tuple in which the first entry is the `ground-truth' label $y^{\ast}$, and the second entry is the predicted label $\hat{y}$. With the set of possible classification outcomes defined, one must specify their relative preference for each outcome. These preferences can be encoded via a \textit{cost function} $C : Y \times Y \rightarrow \mathbb{R}^{K \times K}$. In practice, $C$ can be expressed as a $K \times K$ matrix whose $i,j^{\text{th}}$ entry, $c_{i,j}$, is (reflective of) the cost in a decision process as a result of assigning label $\hat{y}=j$ to a datapoint whose true label is $y^{\ast}=i$. Here, it is worth noting that the current paper adopts a convention for the cost function such that adverse outcomes (misclassifications) are assigned positive numbers; one could equally construct a \textit{utility function} in which adverse outcomes are assigned negative numbers.

For structural digital twin applications, the cost function $C$ would ideally be specified when conducting an operational evaluation, prior to the implementation of the system. In practice, defining the cost function $C$ could prove not-straightforward. In many situations, the specification would likely require engaging domain experts in an elicitation process in order to glean estimates of the costs associated with taking incorrect actions as a result of misclassifications. While some erroneous actions have definitive costs that can be expressed monetarily (perhaps, for example, unnecessarily sending an engineer to perform a structural inspection), others may be uncertain or ill-defined. Consider, as an example, the SHM scenario in which an undamaged label is assigned erroneously resulting in a missed intervention. In this scenario, it would be difficult to say with certainty what the ultimate consequence of a missed intervention would be as it would be conditional on the current rate of deterioration, as well as the likelihood of correctly identifying the damage in the future, prior to structural failure. For such situations, it may be advisable to represent one's preference in terms of a \textit{risk}, or \textit{expected utility}, rather than a direct cost. The probabilities necessary to define these risks could again be based on an expert's prior belief regarding the likelihood of outcomes, or they could be estimated using physics-based simulation.
 
\subsection{Formulation}

To reflect the desire to reduce the number of misclassifications, the proposed cost-informed approach to dimensionality reduction is based around the notion of separability and as such, follows an intuition similar to that outlined for LDA in Section \ref{sec:background}.

Once again, considering a $N \times D$ data matrix $\mathbf{X}$, the total empirical scatter $\Sigma_e$ can be obtained with Equation (\ref{eq:empirical_scat}). Here, using the total scatter captures both the scatter between classes and the scatter within classes (increasing either one would increase the total scatter), and it can therefore can be used as the denominator in a representation of separability akin to Equation (\ref{eq:separability}). It should be noted here that considering the total empirical scatter relaxes the assumption used in the formulation of LDA presented in the previous section that the scatter within classes is common to all.

Since misclassification costs are defined over pairs of labels, it follows that one should consider the scatter between classes in a pairwise manner. For two classes $i$ and $j$, the pairwise scatter between classes is represented by the weighted average,

\begin{equation}\label{eq:pairwise_scat}
	\Sigma_{i,j} = \frac{1}{n_i + n_j} \Big[ n_i (\mathbf{X}_i - \bar{\mathbf{x}}_{i,j})^{\top} (\mathbf{X}_i - \bar{\mathbf{x}}_{i,j}) + n_j (\mathbf{X}_j - \bar{\mathbf{x}}_{i,j})^{\top} (\mathbf{X}_j - \bar{\mathbf{x}}_{i,j}) \Big] ,
\end{equation}

\noindent where $n_i$ and $n_j$ are the number of datapoints associated with class $i$ and class $j$, respectively; $\mathbf{X}_i$ and $\mathbf{X}_j$ denotes the data associated with classes $i$ and $j$, respectively; and where $\bar{\mathbf{x}}_{i,j}$ denotes the empirical mean, or centroid, of the combined data for both classes $i$ and $j$, i.e.\ $\bar{\mathbf{x}}_{i,j} = \frac{\bar{\mathbf{x}}_i + \bar{\mathbf{x}}_j}{2}$.

Now, recall the definition of direction-dependent separability given in Equation (\ref{eq:separability}). It can be understood from this equation that increasing the between-class scatter $\Sigma_b$ yields higher separability -- this property can be exploited for the purpose of cost-informed dimensionality reduction. Rather than constructing the between-class scatter using the class centroids $\bar{\mathbf{x}}_k$ and the overall centroid $\bar{\mathbf{x}}$ per Equation (\ref{eq:LDA2}), one can construct the between-class scatter by considering instead the scatters between pair of classes given in Equation (\ref{eq:pairwise_scat}). It is here that one can introduce a weighting to increase (or decrease) the contribution of each pairwise scatter matrix  to $\Sigma_b$ according to the cost of misclassification as specified by $C$; doing so yields,

\begin{equation}\label{eq:CISb}
	\Sigma_b = \sum_{i=1}^{K} \sum_{j=1}^{K} c_{i,j}\Sigma_{i,j}.
\end{equation}

 One is now in a position to define a representation of cost-weighted separability,

\begin{equation}
	\psi_C = \frac{\mathbf{u}^{\top} \Sigma_b \mathbf{u}}{\mathbf{u}^{\top} \Sigma_e \mathbf{u}},
\end{equation}

and a corresponding optimisation,

\begin{equation}\label{eq:CIopt}
	\hat{\mathbf{u}} = \underset{\mathbf{u}}{\text{argmax}}\psi_C.
\end{equation}

From Equations (\ref{eq:CISb}) to (\ref{eq:CIopt}), it is apparent that weighting the $\Sigma_{i,j}$ according to the corresponding misclassification cost $c_{i,j}$, in essence, artificially inflates the scatter between classes $i$ and $j$ and thus its contribution to $\Sigma_b$ for more concerning types of misclassification. Following some careful thought, one can subsequently come to the realisation that the principal directions associated with the pairwise scatter matrices $\Sigma_{i,j}$ will then contribute more to $\hat{\mathbf{u}}$ if the misclassification cost $c_{i,j}$ is larger; thus, the directions in which classes with high cost of misclassification are separable will be prioritised in the optimisation.

As with PCA and LDA, the optimisation can be accomplished via an eigendecomposition,

\begin{equation}
	\Sigma_r = V \Lambda V^{\top}
\end{equation} 

\noindent where $\Sigma_r = \Sigma_e^{-1} \Sigma_b$. In this formulation, the eigenvalues are related to a cost-weighted separability and the corresponding eigenvectors preserve cost-weighted separability. Once again, the data can be transformed to the new coordinate system via Equation (\ref{eq:proj}) and dimensionality can be reduced by removing columns.

As with LDA, the approach detailed above requires class labels to be associated with the data and as such it can be considered a supervised learning approach.

The current section has detailed a proposed approach for conducting cost-informed dimensionality reduction motivated by the need to account for non-uniform and asymmetric misclassification costs that are prevalent in structural asset management decision processes. The following section demonstrates the proposed approach using a visual case study.

\section{A Visual Case Study}\label{sec:CaseStudy}

To compare the proposed cost-informed dimensionality reduction technique with standard approaches (PCA and LDA), a synthetic dataset was constructed. The dataset consists of a three-dimensional ($D=3$) input space $\mathbf{x}_t = \{ x_1 , x_2, x_3 \}$ and a nine-class label space $y \in \{ 0,1,2,3,4,5,6,7,8 \}$. While a three-dimensional input space may not be particularly troubling from a modelling standpoint, this dimensionality allows one to visualise both the original and projected spaces, facilitating verification that the dimensionality reduction techniques are behaving as expected, and potentially granting additional insights.

\begin{figure}[h!]
	\centering
	\scalebox{0.6}{
		\includegraphics{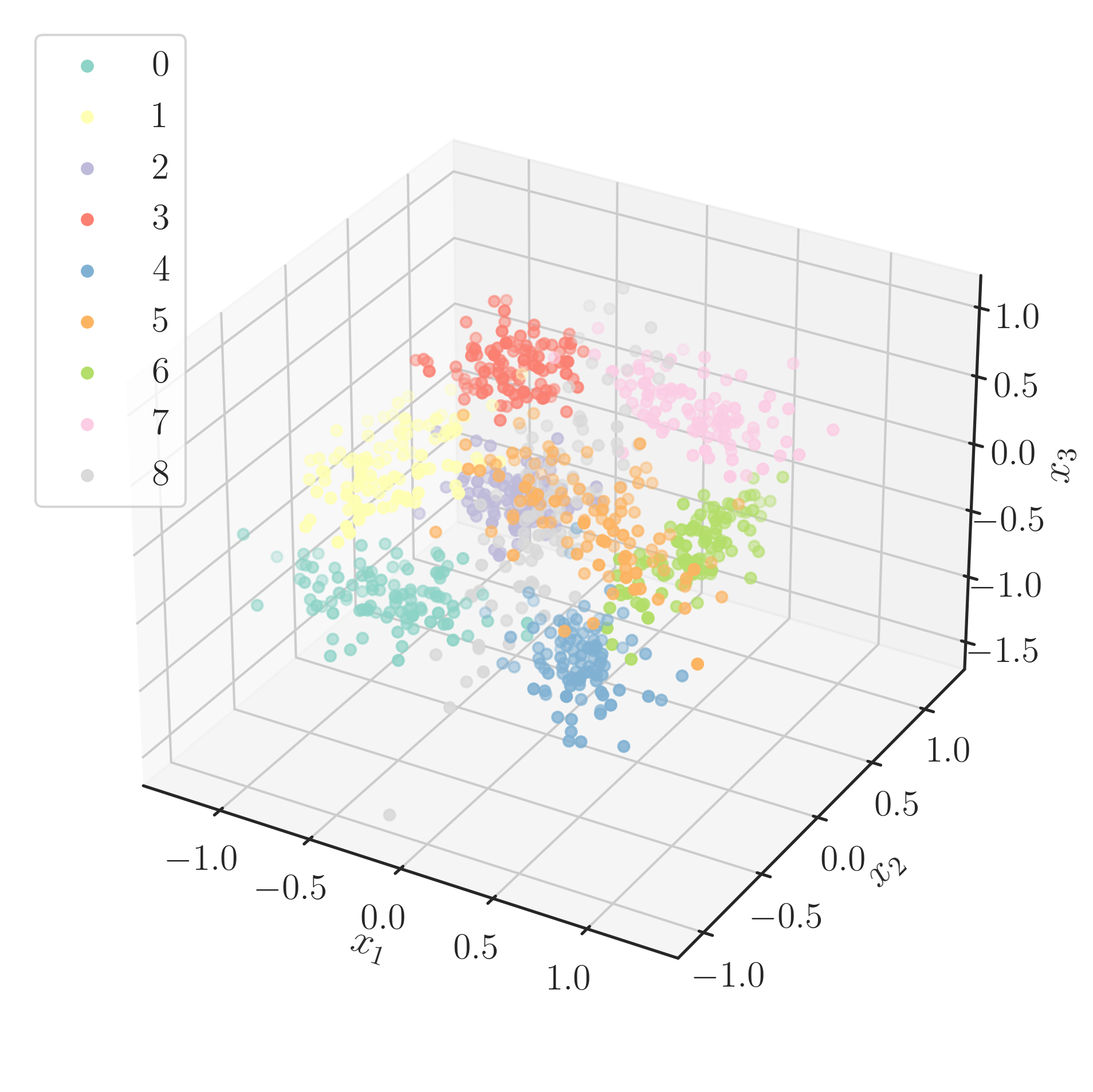}
	}
	\caption{Synthetic dataset in its original three dimensions.}
	\label{fig:data_3d}
\end{figure}

A generative model was constructed by defining Gaussian distributions with means $\bm{\mu}_k$ at the vertices of a unit cube with a geometric centre at the origin $\{0,0,0\}$; a ninth class was included at this geometric centre. Covariances for each of the classes were sampled at random from an inverse-Wishart distribution with the following parametrisation,

\begin{equation}
	\Sigma_k \sim \mathcal{W}^{-1}(\Psi=0.15\cdot\mathbb{I}_3,\nu=8)
\end{equation}

\noindent where $\Sigma_k$ denotes the $3\times3$ covariance matrix for class $k$, $\Psi$ denotes the inverse-Wishart scale parameter, $\mathbb{I}_3$ denotes the $3 \times 3$ identity matrix, and $\nu$ denotes the inverse-Wishart degrees-of-freedom parameter. From this generative model, 100 datapoints were sampled for each class. A three-dimensional visualisation for one realisation from this generative model is provided in Figure \ref{fig:data_3d}. 

\begin{figure}[h!]
	\centering
	\begin{subfigure}[b]{.45\linewidth}
		\includegraphics[width=\linewidth]{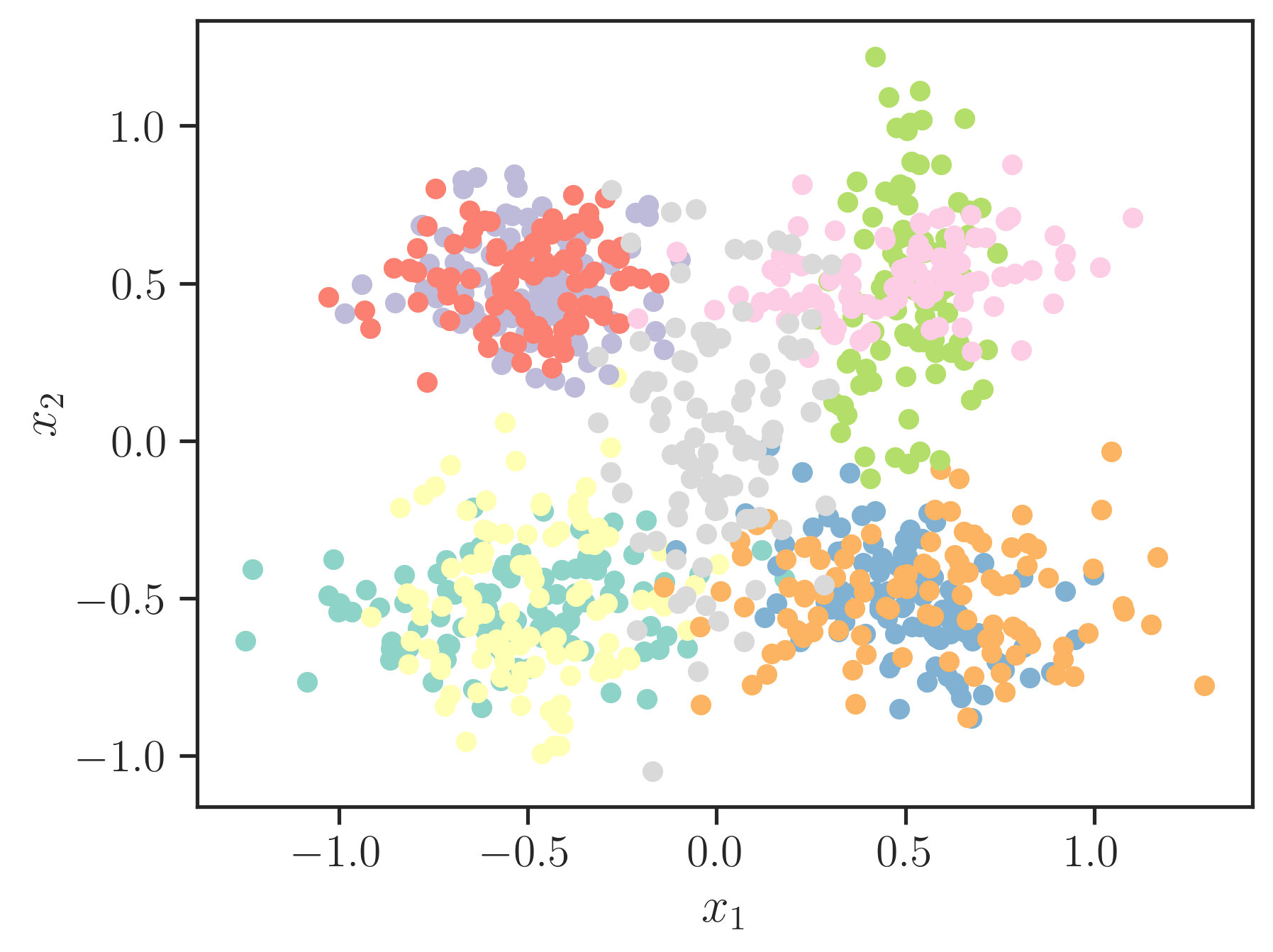}
		\setcounter{subfigure}{0}%
		\caption{}\label{fig:orig12}
	\end{subfigure}
	
	\begin{subfigure}[b]{.45\linewidth}
		\includegraphics[width=\linewidth]{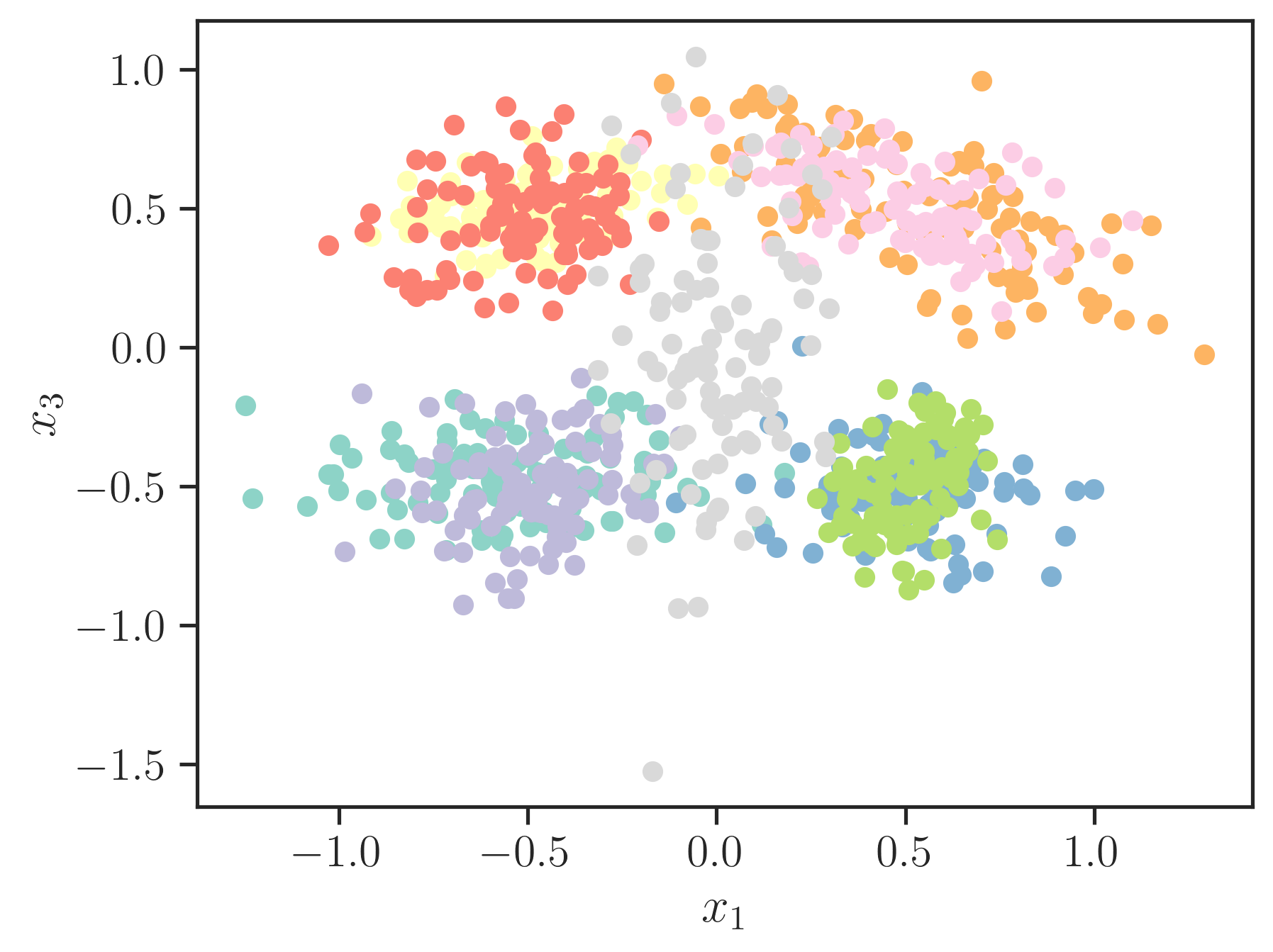}
		\caption{}\label{fig:orig13}
	\end{subfigure}
	\begin{subfigure}[b]{.45\linewidth}
		\includegraphics[width=\linewidth]{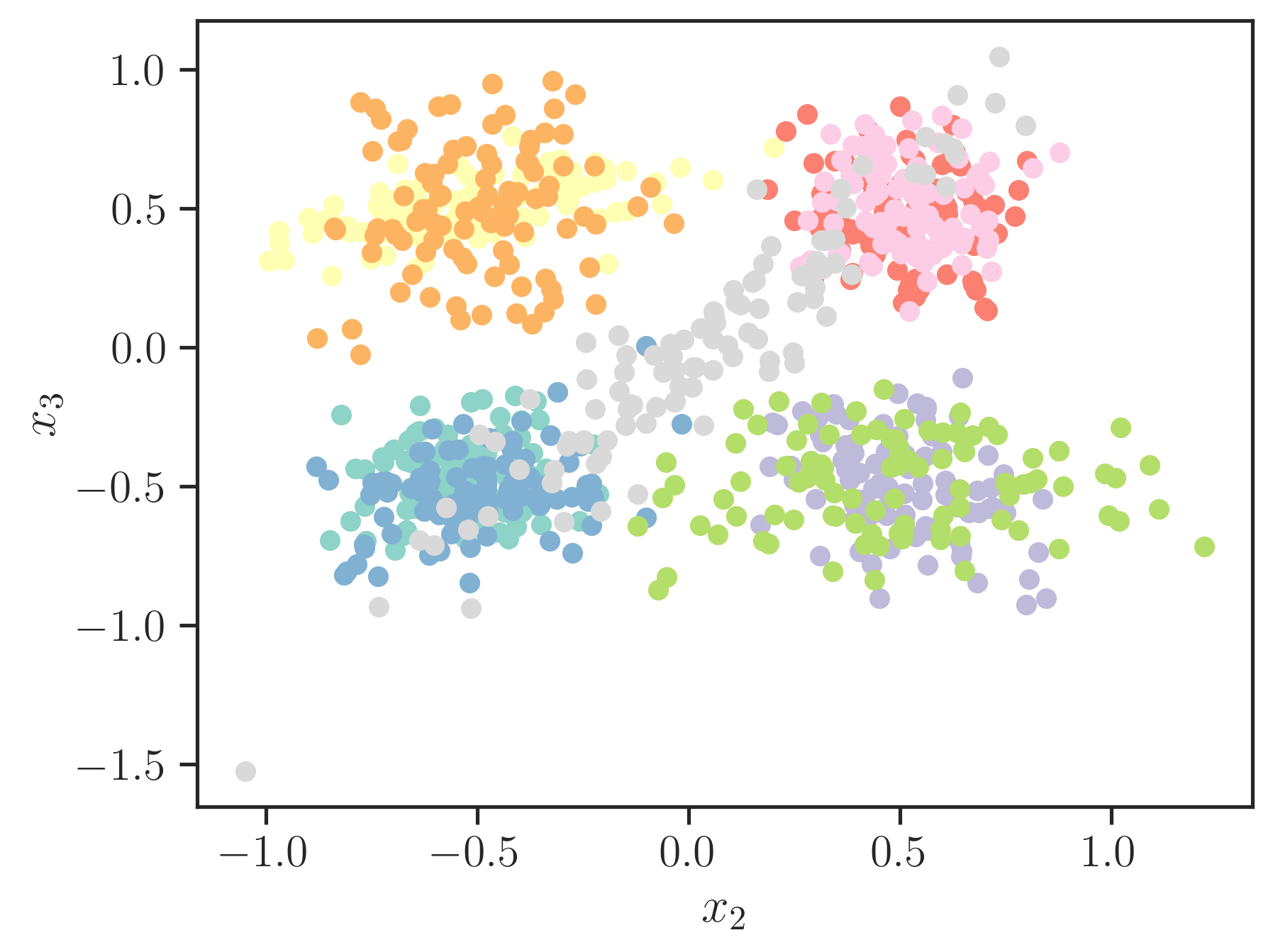}
		\caption{}\label{fig:orig23}
	\end{subfigure}
	\caption{Possible two-dimensional projections of the dataset retaining the original features $x_1$, $x_2$, and $x_3$. The combinations shown are as follows (a) $\{x_1,x_2\}$, (b) $\{x_1,x_3\}$, and (c) $\{x_2,x_3\}$.}
	\label{fig:original_proj}
\end{figure}

In the original three dimensions, the classes are generally well-separated. However, this dataset possesses the characteristic that reducing to any two of the original features will result in some loss of separability between classes. This property can be visualised via Figure \ref{fig:original_proj}; it can be seen that in each projection, each of the clusters on the vertices of the cube coincide with another class.

A classification task can be established for the data set with the goal being to discriminate between the nine classes. For the current case study, a $K$-nearest neighbours classifier is used with $K=5$ \cite{Cunningham2021knn}.

While the datasets considered in this case study are entirely synthetic, for the purposes of studying the behaviour of the proposed dimensionality reduction technique, one can imagine that the class labels correspond to latent states pertinent to an asset management decision process for a structure in operation. Accordingly, misclassification costs can be assigned to the possible classification outcomes; the cost function used in the current case study is shown in Table \ref{tab:costs}.

\begin{table}[h!]
	\centering
	\caption{Cost function $C$.}
	\scalebox{0.8}{
	\begin{tblr}{
			colspec={cQ[c,m,.75cm]Q[c,m,.75cm]Q[c,m,.75cm]Q[c,m,.75cm]Q[c,m,.75cm]Q[c,m,.75cm]Q[c,m,.75cm]Q[c,m,.75cm]Q[c,m,.75cm]Q[c,m,.75cm]}, 
			row{1-2} = {font=\bfseries},
			column{1-2} = {font=\bfseries},
			vline{2} = {3-11}{solid},
			vline{3-12} = {2-11}{solid},
			abovesep=4pt}
		& &\SetCell[c=9]{c}{\bfseries Predicted Label}\\
		\cline{3-11}
		& & 0 & 1&2 &3 &4 &5 &6 &7 &8\\
		\cline{2-11}
		\SetCell[r=9]{c}True label& 0 &  0 & 1 & 10 & 1 & 1 & 1 & 1 & 1 & 1\\
		\cline{2-11}
		& 1 & 1 & 0 & 1 & 1 & 1 & 1 & 1 & 1 & 10\\
		\cline{2-11}
		& 2 & 50 & 1 & 0 & 1 & 1 & 1 & 1 & 1 & 1\\
		\cline{2-11}
		& 3 & 1 & 1 & 1 & 0 & 1 & 1 & 1 & 10 & 1\\
		\cline{2-11}
		& 4 & 1 & 1 & 1 & 1 & 0 & 1 & 10 & 1 & 1\\
		\cline{2-11}
		& 5 & 1 & 1 & 1 & 1 & 1 & 0 & 1 & 1 & 1\\
		\cline{2-11}
		& 6 & 1 & 1 & 1 & 1 & 50 & 1 & 0 & 1 & 1\\
		\cline{2-11}
		& 7 & 1 & 1 & 1 & 50 & 1 & 1 & 1 & 0 & 25\\
		\cline{2-11}
		& 8 & 1 & 50 & 1 & 1 & 1 & 1 & 1 & 25 & 0\\
		\cline{2-11}

	\end{tblr}
}
\label{tab:costs}
\end{table}

While the cost function in Table \ref{tab:costs} is somewhat arbitrary, the misclassification costs were defined so as to reflect the non-uniformity and asymmetry prevalent in structural asset management decision processes. Correct classifications were given a cost of zero and, for the most part, misclassifications were assigned a unitary nominal cost. Three misclassifications, (2,0), (6,4), (7,3), and (8,1), were assigned large misclassification costs of 50. Considering an SHM decision process as an example, misclassifications with associated costs such as these could be interpreted as type-II errors in which damage has been incorrectly labelled as undamaged meaning some critical intervention has been missed, thus jeopardising structural integrity and resulting in high-cost incurred. The reverse of those misclassifications, (0,2), (4,6), (3,7), and (1,8), were assigned a smaller cost of 10. In the SHM example, misclassifications with costs such as these could correspond to situations in which an inspection is carried out unnecessarily. Finally, (7,8) and (8,7) were assigned symmetric costs of 25. In an SHM context, misclassifications with costs such as these could correspond to confusion between two damage states, e.g.\ damage has been identified but in the wrong location, which could result in additional time/money spent doing inspections or with the structure out of operation.

With the dataset, classification task, and cost function established, the performance of the cost-informed approach to dimensionality reduction can be assessed.

\subsection{Results}

\begin{figure}[h!]
	\centering
	\begin{subfigure}[b]{.45\linewidth}
		\includegraphics[width=\linewidth]{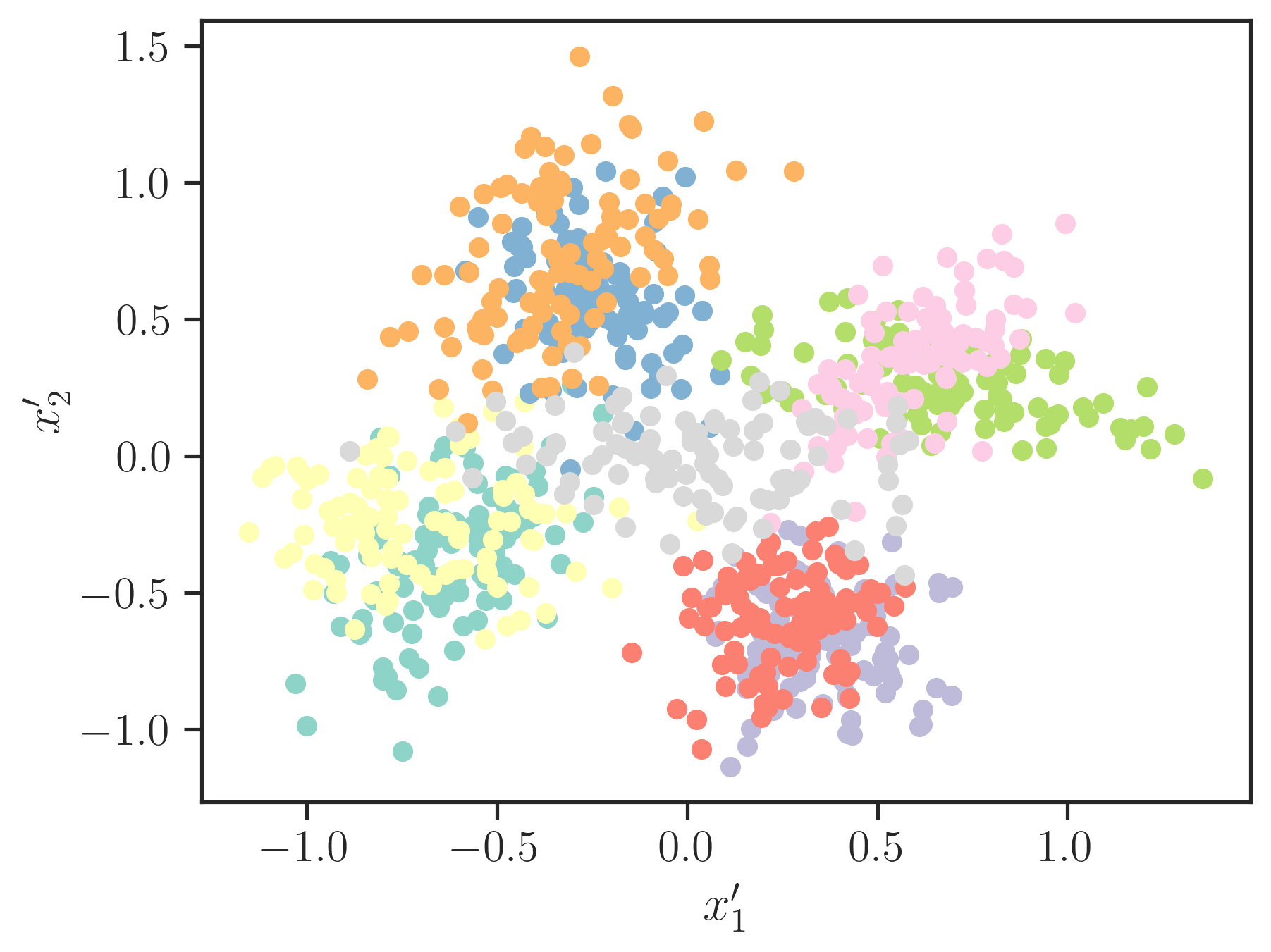}
		\setcounter{subfigure}{0}%
		\caption{Cost-informed}\label{fig:proj_clda}
	\end{subfigure}
	
	\begin{subfigure}[b]{.45\linewidth}
		\includegraphics[width=\linewidth]{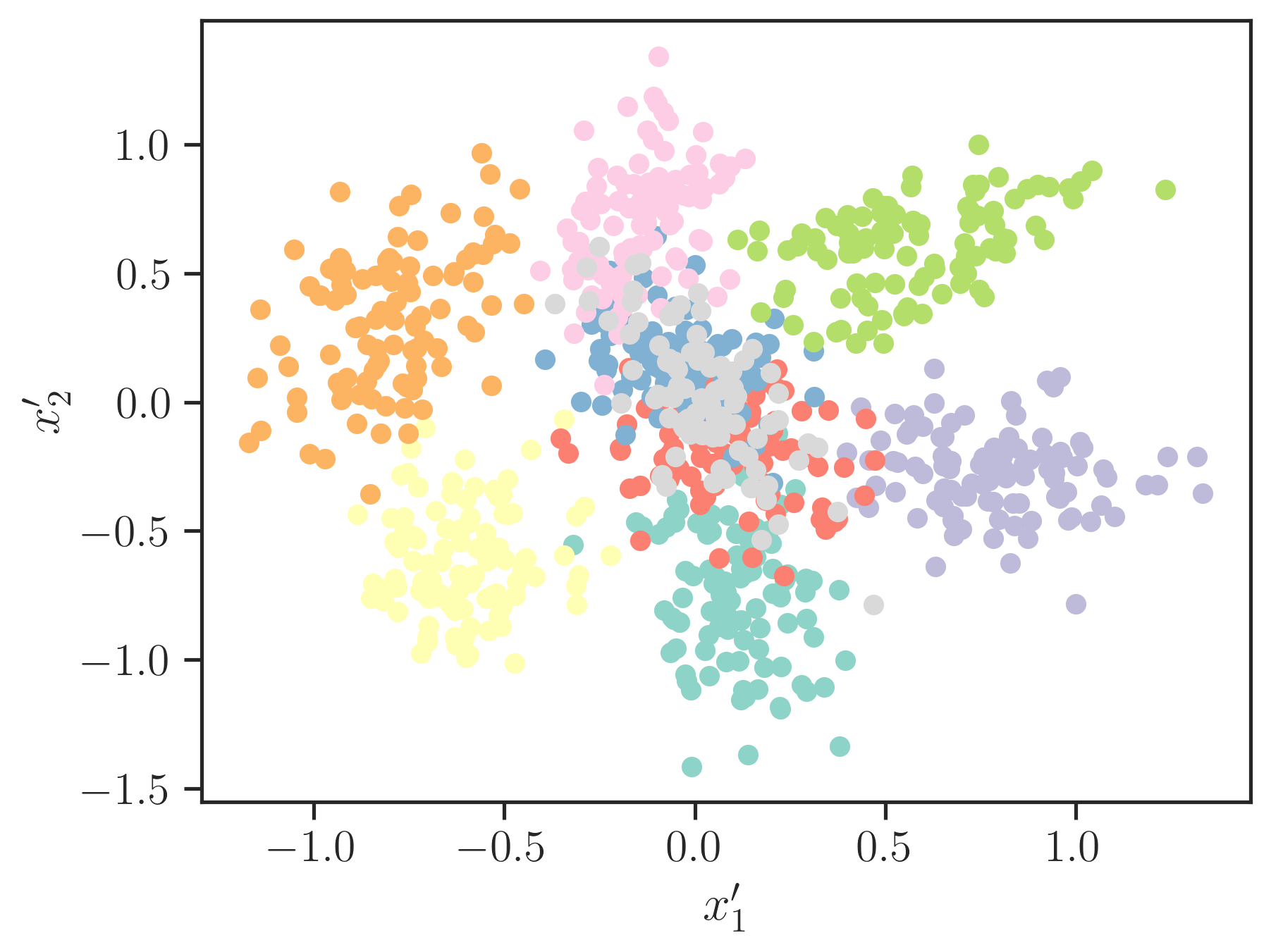}
		\caption{LDA}\label{fig:proj_lda}
	\end{subfigure}
	\begin{subfigure}[b]{.45\linewidth}
		\includegraphics[width=\linewidth]{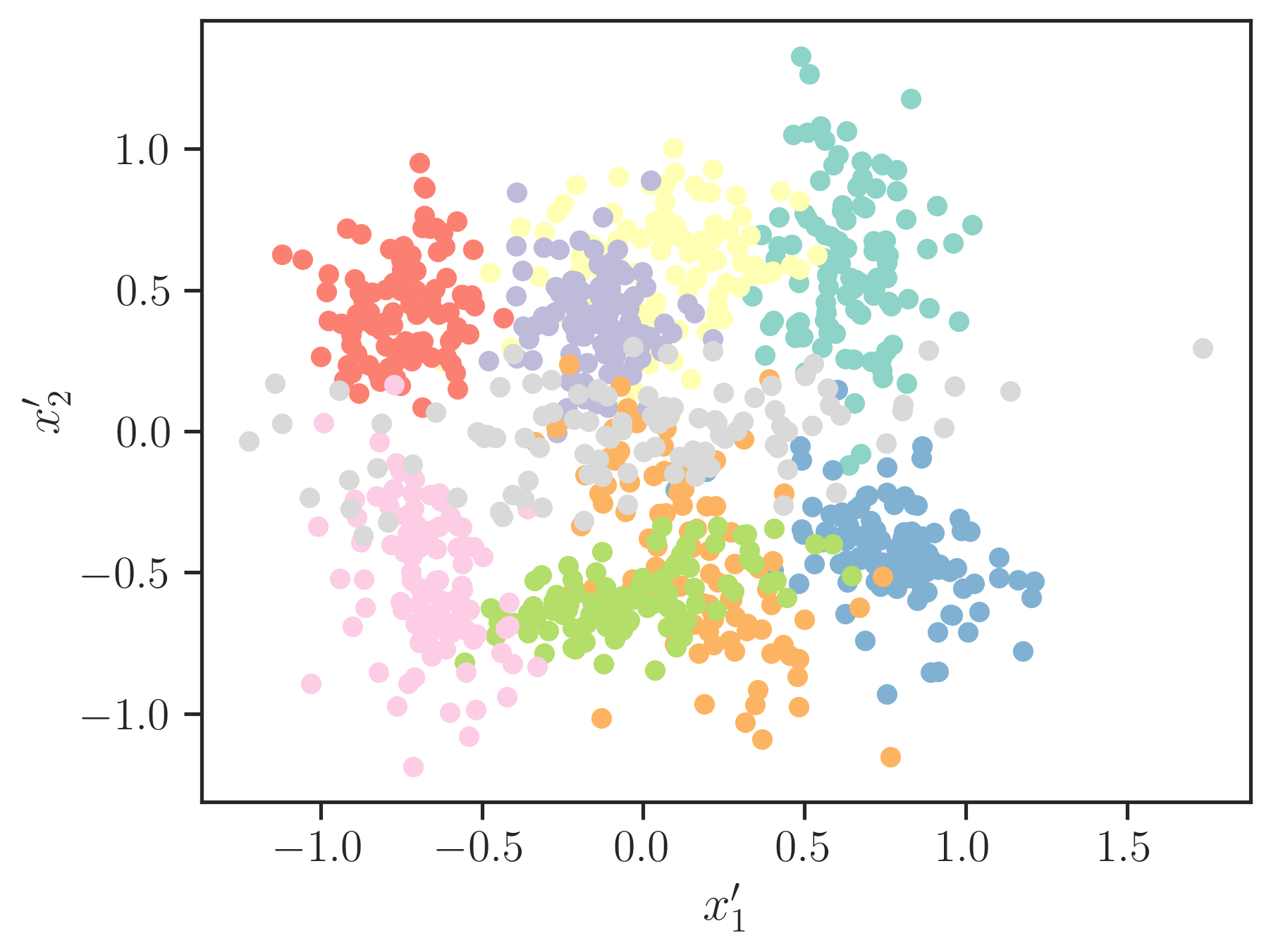}
		\caption{PCA}\label{fig:proj_pca}
	\end{subfigure}
	\caption{Two-dimensional projections of the dataset, transformed using the first two eigenvectors obtained via (a) cost-informed dimensionality reduction, (b) LDA, and (c) PCA.}
	\label{fig:proj}
\end{figure}

To assess the performance of cost-informed dimensionality reduction, the technique was compared to PCA and LDA. The two-dimensional projections obtained for one of the realisations of the generative model described above is shown in Figure \ref{fig:proj}.

Figures \ref{fig:proj_clda}, \ref{fig:proj_lda}, and \ref{fig:proj_pca} show the 2-dimensional projections for cost-informed dimensionality reduction, LDA, and PCA, respectively. It can be seen from Figure \ref{fig:proj_clda}, that the class pairs specified to have high cost in the cost function $C$ are fairly well-separated; specifically, Classes 0 and 2 (dark green and purple), Classes 1 and 8 (yellow and grey), Classes 3 and 7 (red and pink), Classes 4 and 6 (blue and light green), and Classes 7 and 8 (pink and grey). It can also be seen that outside of these class combinations, some separability has been sacrificed. Overall, this result is promising and suggests that the proposed approach to cost-informed dimensionality reduction is performing as intended. From Figures \ref{fig:proj_lda} and \ref{fig:proj_pca}, it can be seen that LDA and PCA are performing as expected. For LDA (Figure \ref{fig:proj_lda}), one can see that six of the classes centred on the vertices of the cube have been separated; however, the two remaining vertex classes, and the centre class have poor separability. Here, the plane onto which the data are projected roughly coincides with the body diagonal of the cube -- this plane maximises the total distance of the vertices from the centre and therefore yields a high degree of inter-class separability. For PCA (Figure \ref{fig:proj_pca}), it can be seen the the first principle component captures the large variance associated with Class 8 (grey), in addition to the variance associated with the separation of classes in the direction of a face diagonal. The second component, orthogonal to the first, then separates two faces of the cube; thus capturing additional variance.

To assess the performance of the proposed approach to dimensionality reduction in terms of the total cost of misclassifications, 500 datasets were generated as randomly-sampled realisations of the generative model described in Section \ref{sec:CaseStudy}. Each dataset was divided into two such that there were 50 points for each class in each of the halves, thus forming training and test sets.

\begin{figure}[h!]
	\centering
	\scalebox{0.8}{
		\includegraphics{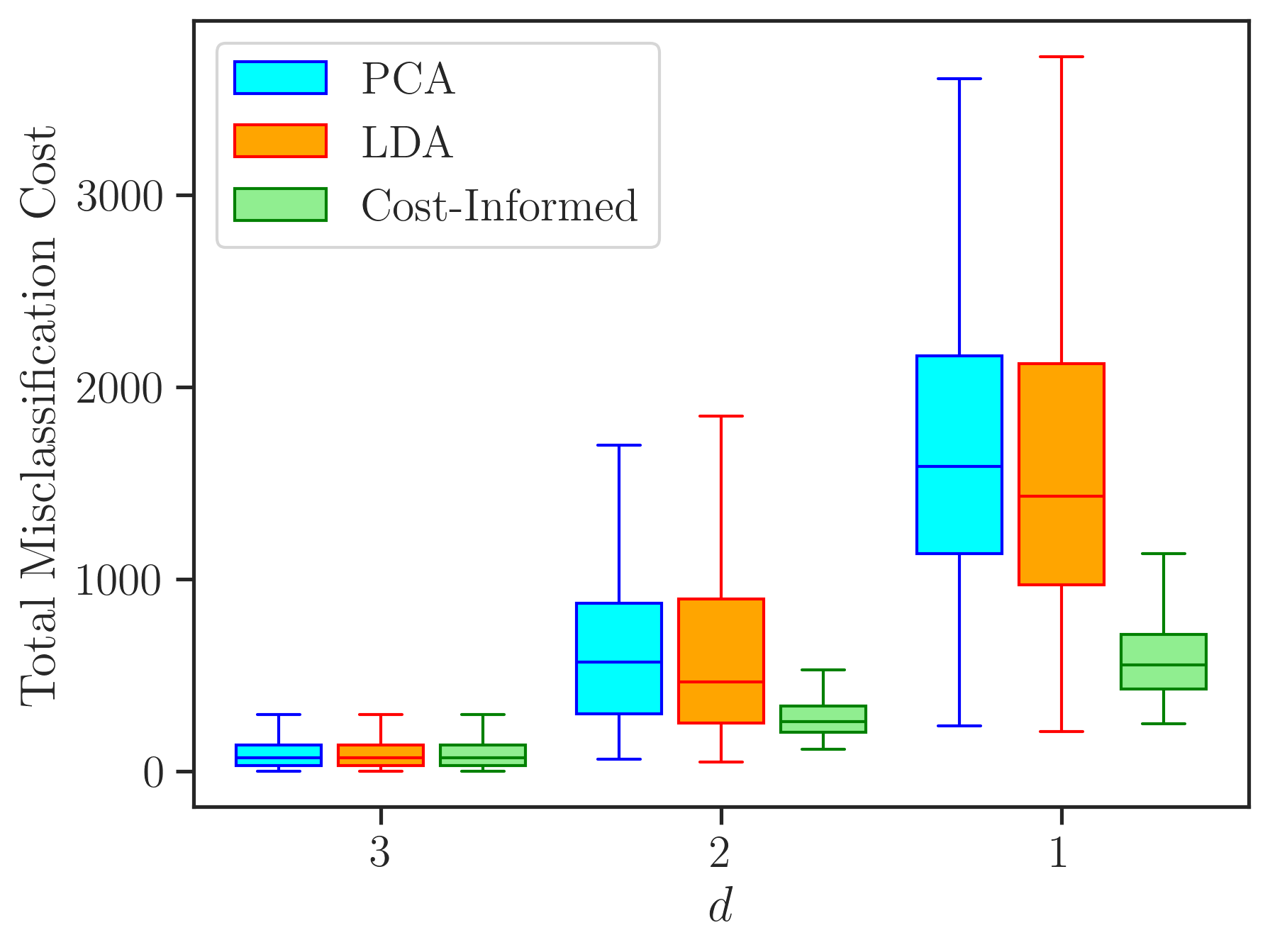}
	}
	\caption{Box plots representing the distributions of total misclassification costs for models trained on data projected using PCA, LDA, and cost-informed LDA, as the number of dimensions is reduced.}
	\label{fig:total_costs}
\end{figure}

The training data were used to learn projections using the cost-informed approach, LDA, and PCA. Then, for each dimensionality, the transformed training data were used to learn a $K$-nearest neighbours classifier. Predictions were then made for the test datasets and for each, the classification performance was summarised in a confusion matrix. The total cost of misclassifications for each test set was evaluated by multiplying the confusion matrices with C and computing the total sum. This process yielded 500 total misclassification costs for each of the three dimensionality reduction techniques and for each of the three possible dimensionalities. The distributions of these costs are represented as box plots in Figure \ref{fig:total_costs}.

It can be seen from Figure \ref{fig:total_costs} that for three dimensions, all projection perform equally, with low misclassification costs across the board. This result is expected as the transformation are comprised of unit vectors and essentially represent rotations of the original space in which the classes were generally separable. When the dimensionality is reduced to two, PCA and LDA start performing markedly worse; both in terms of the median cost and the range. While there is some degradation in performance for the cost-informed approach, it is to a much lesser degree than for the other two techniques. This trend continues as the dimensionality is reduced to one. Once again, PCA and LDA have large increases in total misclassification cost, whereas the cost-informed approach has a much less severe increase. It can be seen in Figure \ref{fig:total_costs} that, for the current case study, PCA and LDA perform quite similarly; this result could possibly be explained by the between-class scatter being the dominant contribution to the total scatter when compared to within-class scatter -- this condition would result in PCA and LDA finding similar projections in many instances.

Overall, these results are very positive as they clearly indicate that, for the presented case study, the proposed approach to cost-informed dimensionality reduction is highly effective in retaining value in classifier-driven decision-making processes when performing dimensionality reduction. These results have significance because they suggest that such approaches to dimensionality reduction can be employed when developing structural digital twin technologies to facilitate meeting key design criteria and constraints, such as available computational resources, in a cost-effective manner.

To summarise, using the synthetic case study presented in Section \ref{sec:CaseStudy}, the performance of the proposed approach to cost-informed dimensionality reduction was assessed and compared to LDA and PCA. It was shown that, as dimensionality was reduced, the feature spaces learned using the cost-informed approach consistently yielded low misclassification costs over test datasets. These results indicate that cost-informed approaches to dimensionality reduction can be used to improve the performance of structural digital twin technologies in decision-support applications.

\section{Discussion}\label{sec:disc}

Although it was demonstrated in the previous section that cost-informed dimensionality reduction can be highly effective in the context of decision-support, the proposed approach has limitations nonetheless. One of these limitations is that the approach, as presented, is limited to linear transformations. There are many applications where data are not linearly separable in $D$ dimensions; thus, it may be advantageous to formulate nonlinear approaches to cost-informed dimensionality reduction \cite{Lawrence2012unifying}. One possible way in which this could be achieved is to first map the data using a nonlinear kernel before applying the cost-informed dimensionality reduction. Further investigation into this method, and other approaches, is left for future work.

Another limitation of the proposed approach is that it is supervised, requiring fully-labelled data. Fully-labelled dataset are commonly unavailable in structural asset management applications as they can be prohibitively expensive, or otherwise impractical, to obtain. For this reason, it could be beneficial to develop a semi-supervised approach to learning the transformation required to perform the dimensionality reduction -- a sensible first step to achieving this would be to extend the approach to a probabilistic formulation. Again, this avenue of research is left as future work.

One area, yet unexplored, which could see substantial benefits from the application of cost-based dimensionality reduction is active learning in the context of structural asset management. Recent works have developed decision-theoretic approaches to active learning in which structural inspections are mandated according to value of information \cite{Hughes2022,Hughes2022robust}. Increasing the separability between classes warranting differing courses of action will increase the confidence in predictions and diminish the regions of the feature space with high value of information. This effect, in turn, will result in fewer structural inspections being made thereby saving further resources beyond the reduction in misclassifications.

\section{Conclusions}\label{sec:conc}

Statistical classifiers are a key technology underpinning many decision support technologies including structural digital twins. Misclassifications correspond to suboptimal actions being taken and often translate to a real cost incurred to the end-user. For structural asset management decision processes, misclassifications can often correspond to costs that are asymmetric and non-uniform.

 Dimensionality reduction is a critical aspect of developing effective statistical classifiers. In particular, it can help mitigate some of the issues associated with the `curse of dimensionality' such as inhibited predictive performance, and it can reduce the computational complexity of prediction. Many commonly used dimensionality reduction techniques are formulated from an information-theoretic perspective and therefore treat all types of misclassification with equal priority -- this assumption can render such approaches suboptimal for decision-support applications. 

The current paper proposes a decision-theoretic approach to dimensionality reduction that accounts for asymmetric and non-uniform misclassification costs. In this approach, bases that preserve separability between classes with high cost of misclassification are prioritised by weighting them in an eigendecomposition via a cost function. It was demonstrated using a synthetic case study that this cost-informed approach can be highly effective at reducing the costs incurred as a result of misclassifications. Overall, the work presented motivates the further development of cost-informed and decision-theoretic approaches to dimensionality reduction for application in structural asset management technologies such as digital twins. The cost savings achieved by the use of such technologies will provide impetus to the adoption of structural digital twins which are needed for critical tasks such as infrastructure lifetime extension.


\section*{Acknowledgements}

The authors would like to gratefully acknowledge the support of the UK Engineering and Physical Sciences Research Council (EPSRC) via grant reference EP/W005816/1. For the purposes of open access, the authors have applied a Creative Commons Attribution (CC BY) license to any Author Accepted Manuscript version arising.



\bibliography{ISMA_2022}


	%
	%
	%


%

\end{document}